\documentclass{article}
\usepackage{ijcai24}

\usepackage{times}
\usepackage{soul}
\usepackage{url}
\usepackage[hidelinks]{hyperref}
\usepackage[utf8]{inputenc}
\usepackage[small]{caption}
\usepackage{graphicx}
\usepackage{amsmath}
\usepackage{amsthm}
\usepackage{booktabs}
\usepackage{algorithm}
\usepackage{algorithmic}
\usepackage[switch]{lineno}
\usepackage{color}

\usepackage{subfigure}
\usepackage{amssymb}
\usepackage{framed,multirow}
\usepackage{bm}
\usepackage{pifont}
\usepackage{makecell}

\usepackage[T1]{fontenc}

\title{Learning Spatial Similarity Distribution for Few-shot Object Counting}

\author{
Yuanwu Xu
\and
Feifan Song\and
Haofeng Zhang$^*$
\affiliations
School of Artificial Intelligence,
Nanjing University of Science and Technology
\emails
\{xuyuanwu, sff, zhanghf\}@njust.edu.cn
}

\begin{document}

\maketitle
\def\thefootnote{$*$}
\footnotetext{Corresponding author.}
\def\thefootnote{\arabic{footnote}}
\begin{abstract}
    Few-shot object counting aims to count the number of objects in a query image that belong to the same class as the given exemplar images. Existing methods compute the similarity between the query image and exemplars in the 2D spatial domain and perform regression to obtain the counting number. However, these methods overlook the rich information about the spatial distribution of similarity on the exemplar images, leading to significant impact on matching accuracy. To address this issue, we propose a network learning Spatial Similarity Distribution (SSD) for few-shot object counting, which preserves the spatial structure of exemplar features and calculates a 4D similarity pyramid point-to-point between the query features and exemplar features, capturing the complete distribution information for each point in the 4D similarity space. We propose a Similarity Learning Module (SLM) which applies the efficient center-pivot 4D convolutions on the similarity pyramid to map different similarity distributions to distinct predicted density values, thereby obtaining accurate count. Furthermore, we also introduce a Feature Cross Enhancement (FCE) module that enhances query and exemplar features mutually to improve the accuracy of feature matching. Our approach outperforms state-of-the-art methods on multiple datasets, including FSC-147 and CARPK. Code is available at \url{https://github.com/CBalance/SSD}.
\end{abstract}

\section{Introduction}
\label{sec:intro}
Visual object counting aims at counting how many times a certain object occurs in the query image, which has received growing attention in the past years. Existing methods often focus on specific domains, such as crowd counting \cite{shu2022crowd,wang2020nwpu,abousamra2021localization}, animal counting \cite{arteta2016counting}, or car counting \cite{hsieh2017drone}. These methods typically rely on large amounts of  data to train accurate counting models. Furthermore, they are limited to counting objects of specific categories and cannot generalize well to novel categories.

To overcome these limitations, a recent approach called Few-shot Object Counting (FSC) has been introduced and gained great attention with the emergence of a dataset \cite{ranjan2021learning}. FSC addresses the challenge of counting objects from arbitrary categories using only a few exemplars. This enables the model to generalize to unseen categories, offering potential for application in various scene categories beyond those encountered during training. By leveraging few exemplars, FSC provides a more flexible and adaptable solution for object counting tasks.

\begin{figure}[!t]
\centering   
{\includegraphics[width=\linewidth]{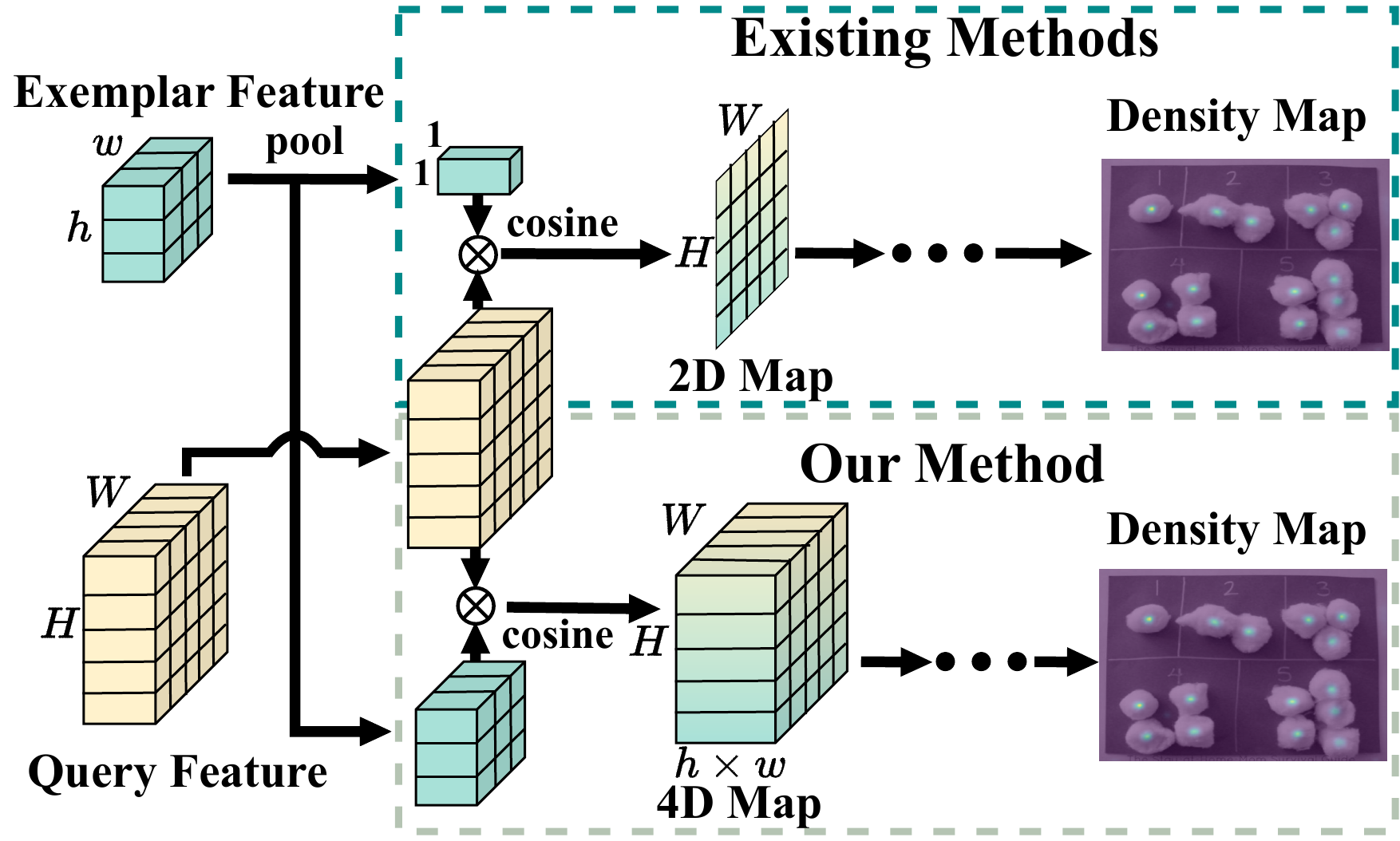}}
\caption{Comparison between existing methods and our method. Compared to the feature similarity computation process in previous methods, our approach preserves the spatial structure of exemplars. Each position is computed with query features, and in the subsequent convolutional regression process, we fully utilize the spatial similarity distribution information between query and exemplar features at a point-to-point level.}
   \label{fig1}
\end{figure}

\begin{figure}
  \centering
    \includegraphics[width=\linewidth]{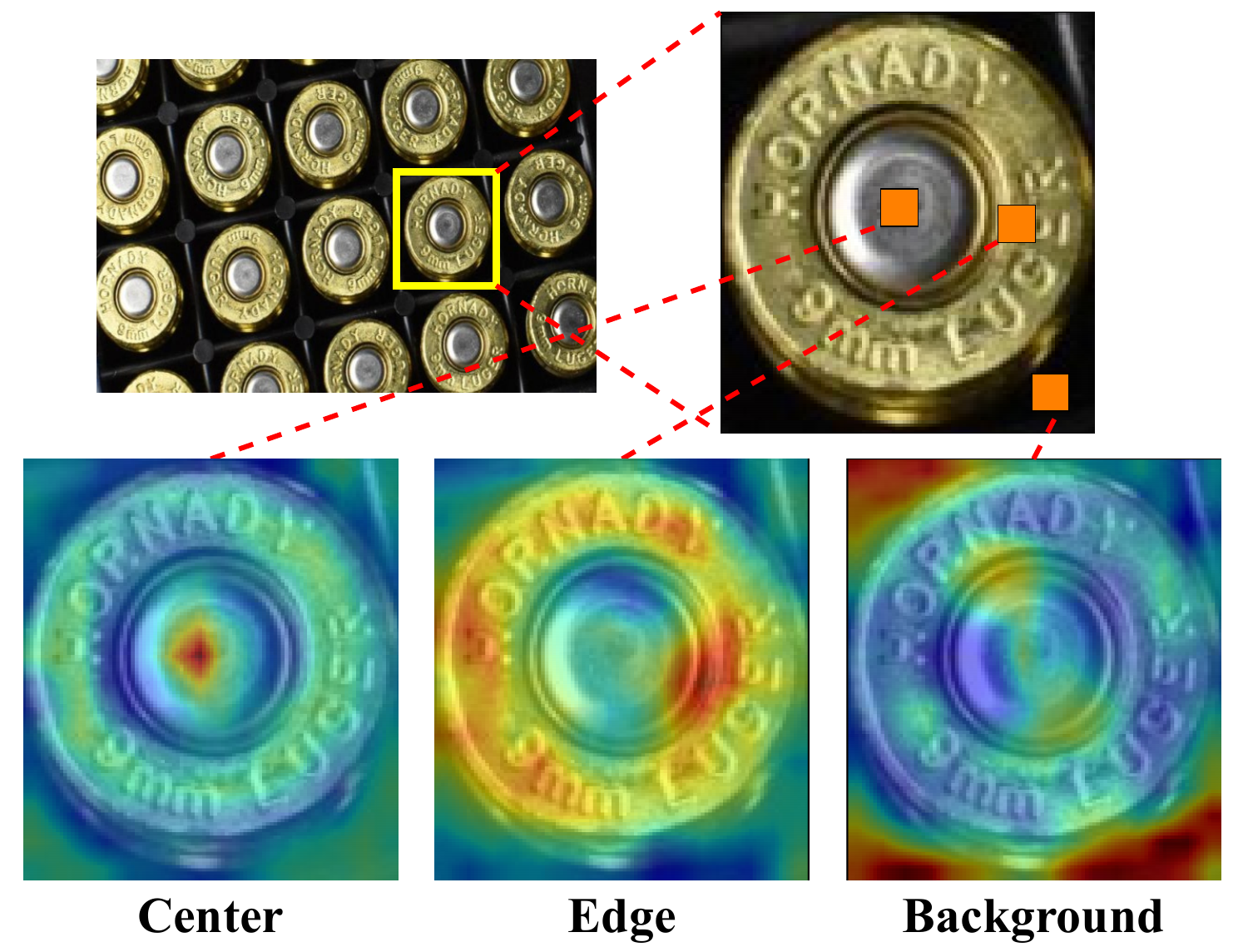}
  \caption{Heatmap depicting the similarity distribution of objects at different positions on the exemplar. }
  \label{fig2}
\end{figure}

As shown in Fig.\ref{fig1}, existing few-shot object counting methods typically follow a general workflow. They first calculate the similarity between query and exemplar features, and then directly regress the similarity matrix or enhance the query features using the similarity matrix and exemplar features before regression. In terms of similarity computation, some methods, as demonstrated in \cite{ranjan2021learning,yang2021class,you2023few,djukic2023low}, employ exemplar features as fixed kernels to perform convolution with query feature. However, in this approach, the distribution of kernel features remains fixed during convolution matching, limiting its adaptability to different sizes and shapes of object features in the query. Another approach, such as the one used in \cite{shi2022represent,Lin_2022_BMVC,liu2022countr}, involves pooling exemplar features to obtain $1\times 1$ feature prototypes, followed by cosine similarity computation between feature vector of each position in the query and these prototypes. This method disregards the distribution information of query and exemplar features, and counting performance becomes dependent on the performance of previous feature extraction and self-attention mechanisms.

To accurately locate the center of an object and generate an appropriate density distribution, we leverage distinct similarity distribution characteristics between each parts, such as object centers, edges and background, when compared to exemplars. Explicitly, as shown in Fig.\ref{fig2}, the similarity distribution of the object center in exemplars gradually diminishes from the central position towards the surrounding regions, while the similarity distribution at the edges exhibits variations across different locations. On the other hand, the background demonstrates generally lower similarity values across all positions except the background area. Taking advantage of these patterns, we propose a novel method that tries to preserve the spatial structure of exemplars during similarity computation, and name it as learning Spatial Similarity Distribution (SSD). Concretely, this method yields a 4D similarity tensor, which allows for flexible extraction of point-to-point similarity distribution information between query and exemplar features using convolution operations in the 4D space. The features obtained through convolution enable precise calculation of density values for each position in the query during regression. In addition, we introduce a Feature Cross Enhancement (FCE) module for query and exemplar features. This employs the similarity matrices as weights to mutually enhance the features, aiming to achieve higher matching accuracy for objects belonging to the given category.

We conduct comprehensive experiments on two renowned public benchmark datasets, \textit{i.e.}, FSC-147 \cite{ranjan2021learning} and CARPK \cite{hsieh2017drone}. The results clearly illustrate that our approach surpasses the performance of current state-of-the-art methods.
Our contributions can be summarized as follows:
\begin{itemize}
\item[$\bullet$]We design a model based on learning the 4D spatial similarity distribution between query and exemplar features in Similarity Learning Module (SLM). This model is capable of obtaining accurate counting results after comprehensive integration of similarity distribution information among point pairs and their surroundings.
\item[$\bullet$]Before calculating the similarity between query and exemplar features, we introduce a Feature Cross Enhancement (FCE) module, which enhances the interaction between them, reducing the distance between the target objects and exemplar features to achieve better matching performance.
\item[$\bullet$]Extensive experiments on large-scale counting benchmarks, such as FSC-147 and CARPK, are conducted and the results demonstrate that our method outperforms the state-of-the-art approaches.
\end{itemize}

\section{Related Work}
\label{sec:related work}
\subsection{Class-Specific Object Counting}\label{CSC}
Class-specific object counting focuses on counting a specific class of objects, such as crowd \cite{stewart2016end,liang2023crowdclip,lin2023optimal,du2023domain}, animals \cite{arteta2016counting}, or cars \cite{hsieh2017drone}. In related methods, the class information can be incorporated into the feature extraction process without additional classification steps. Existing methods can be broadly categorized into detection-based and regression-based approaches.

Detection-based methods detect the positions of objects in an image to perform counting. However, counting accuracy in these methods relies heavily on the performance of the detection process, which introduces errors. This limits the effectiveness of counting tasks in scenarios with densely packed objects. To address this issue, regression-based methods have been proposed to generate a density map, where the sum of the density values represents the predicted object count.

Classic detection-based methods, for example, \cite{stewart2016end} propose a model that decodes an image into a set of people detections, generating distinct detection hypotheses directly from the input image. On the other hand, recent research in regression-based methods, such as \cite{cheng2022rethinking}, utilizes locally connected multivariate Gaussian kernels as replacements for convolution filters. Moreover, a recent work \cite{liang2023crowdclip} proposes knowledge transfer from a vision-language pre-trained model (CLIP) to unsupervised crowd counting tasks, eliminating the need for density map annotation.

\begin{figure*}[t]
  \centering
    \includegraphics[width=0.95\textwidth]{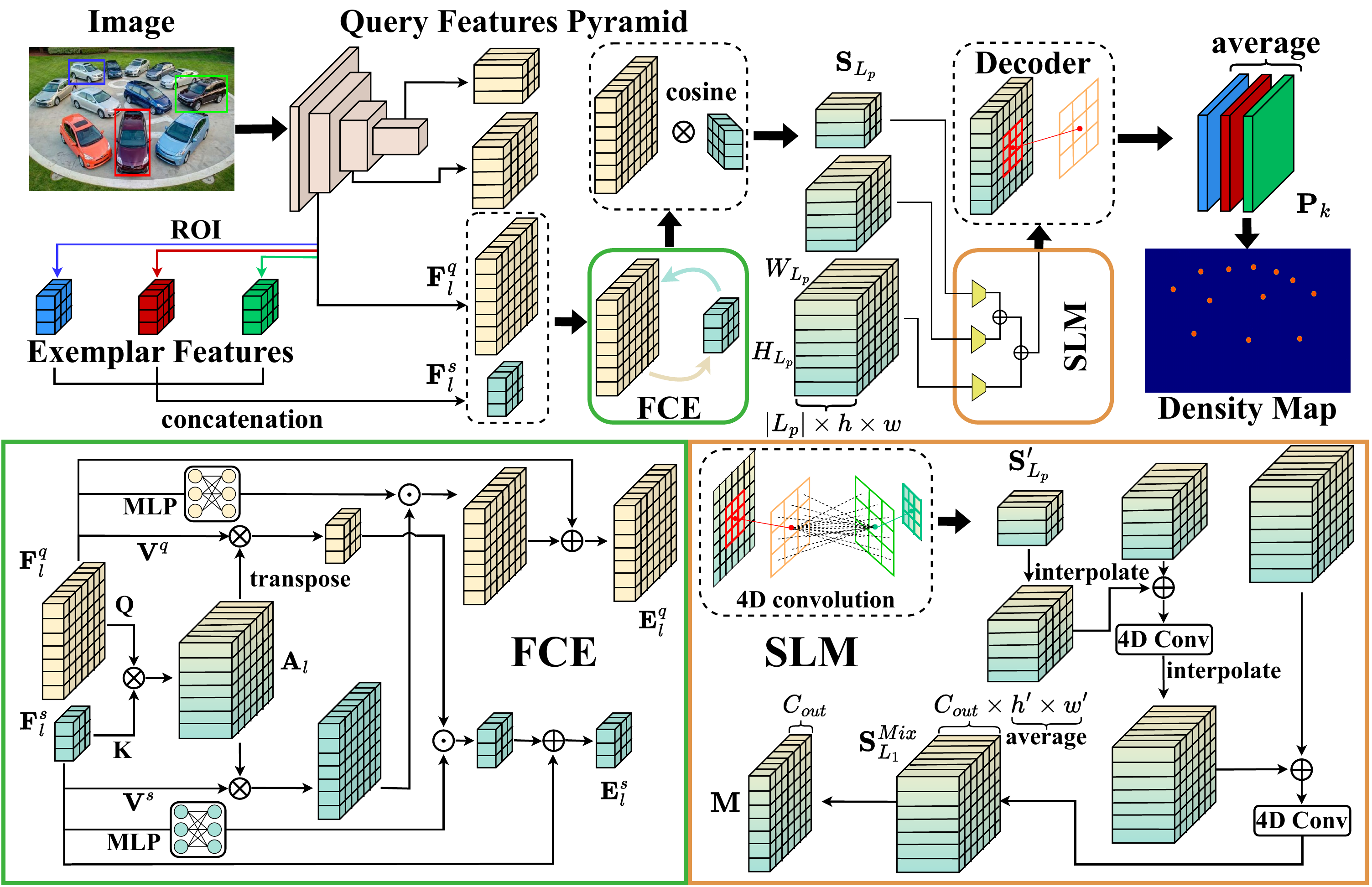}
  \caption{The whole architecture of the proposed SSD framework. }
  \label{fig3}
\end{figure*}

\subsection{Few-shot Object Counting}\label{FSC}
In recent years, few-shot object counting (FSC) has gained significant attention and witnessed a surge of interest. FSC aims to accurately count objects in an image by leveraging only a few exemplars as references. This ability to adapt to unseen categories during the testing phase is a key advantage of FSC.

Several noteworthy methods have been proposed for FSC. GMN \cite{lu2019class} concatenates support features and query features together, and regresses a predicted density map based on this concatenation. In contrast, FamNet \cite{ranjan2021learning} convolves the query image with exemplars used as convolutional kernels, generating multiple similarity maps that provide insights into the comparison results between the query and exemplars. A predicted density map is then regressed from these similarity maps. Another approach, BMNet \cite{shi2022represent}, employs global pooling to transform exemplars into prototypes, and replaces fixed inner product operations with a learnable bilinear similarity metric for comparing exemplar prototypes with query image features. CounTR \cite{liu2022countr} introduced a transformer-based architecture for extracting image features and utilized cross-attention modules for effective feature matching. Recently, LOCA\cite{djukic2023low} is proposed and considers the exemplar shape and appearance properties separately and iteratively adapts these into object prototypes by a new object prototype extraction (OPE) module considering the image-wide features.

\subsection{Generalized Loss}\label{GL}
Generalized loss function \cite{wan2021generalized} is proposed for learning density maps for crowd counting and localization, which is based on unbalanced optimal transport. And \cite{wan2021generalized} prove that both L2 loss and Bayesian loss \cite{ma2019bayesian} are special cases of the generalized loss. The approach proposed in \cite{Lin_2022_BMVC} also utilizes this loss function and introduces a scale-sensitive generalized loss that applies different loss computation methods to object categories of different scales. 

\section{Methodology}\label{sec:Method}
\subsection{Problem Setting}\label{PS}
In few-shot object counting, the dataset is split into base classes $C_{base}$ and novel classes $C_{novel}$, where $C_{base}$ and $C_{novel}$ do not overlap. The remarkable generalization capability of Few-shot Object Counting (FSC) lies in its ability to achieve high performance on the val set and test set, even for categories $C_{novel}$ that have not been encountered during training on $C_{base}$. FSC addresses the task of counting the number of objects of interest present in a query image $\mathbf{X}\in \mathbb{R}^{3\times H\times W}$, with the assistance of $K$ exemplars $\mathbf{Z}$.

\subsection{Overall Architecture}\label{arc}

\begin{figure}[t]
\centering   
{\includegraphics[width=\linewidth]{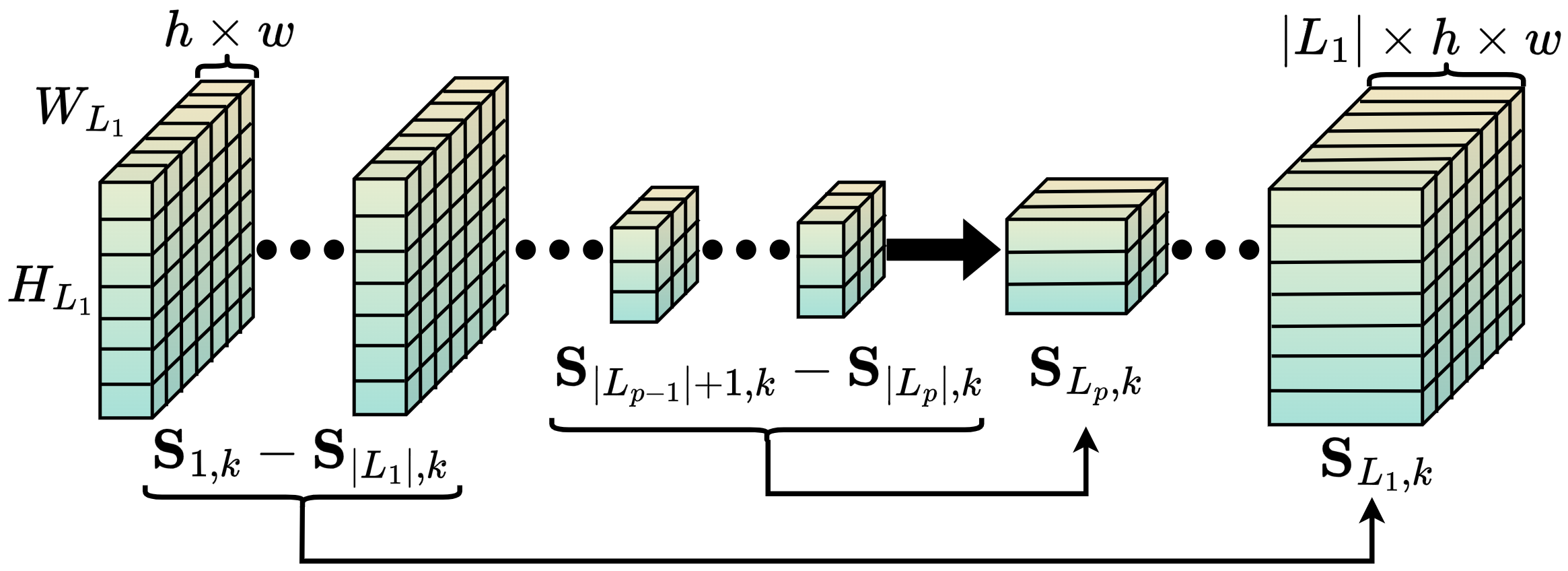}}
\caption{Concatenation of multi-level similarity matrices.}
   \label{fig4}
\end{figure}

As shown in Fig.\ref{fig3}, our entire framework follows the following steps: (1) Feature extraction, (2) Feature Cross Enhancement (FCE), (3) Similarity pyramid calculation, (4) Similarity learning and (5) Regression decoder. Initially, the ResNet-50 \cite{he2016resnet50} feature extractor is used to extract features from the image $\mathbf{X}\in \mathbb{R}^{3\times H\times W}$, with the option of weight freezing (no fine-tuning is performed), and generated pyramid features $\left \{\mathbf{F}_{l}^{q}\right \}_{l=1}^{L}$, where each level feature $\mathbf{F}_{l}^{q}\in \mathbb{R}^{C_{L_{p}}\times H_{L_{p}}\times W_{L_{p}}}\left (l\in L_{p} \right )$. Among all levels, several adjacent levels have features with the same spatial dimensions. All these levels together form a large layer $L_{p} \left (p=1,2,\cdot\cdot\cdot,P \right )$. For each level in the feature pyramid, $K$ exemplar features $\mathbf{F}_{l}^{s}\in \mathbb{R}^{K\times C_{L_{p}}\times h\times w}$ are extracted using the RoIAlign method \cite{he2017mask}. $C_{L_{p}}$ represents the feature channel dimension of the respective layer, while $H_{L_{p}}\times W_{L_{p}}$ and $h\times w$ denote the spatial dimensions of the query feature and exemplar features at that layer. Here we maintain all exemplar features at a uniform size $h\times w$. 

The $K$ exemplar features are input into the FCE module, along with the query feature at the same level, resulting in enhanced features $\mathbf{E}_{l}^{q}$ and $\mathbf{E}_{l}^{s}$ with the same dimensions as $\mathbf{F}_{l}^{q}$ and $\mathbf{F}_{l}^{s}$. We partition the $K$ exemplars, with each exemplar corresponding to a set of feature pyramid combinations $\left \{\mathbf{E}_{l}^{q},\mathbf{E}_{l,k}^{s}\right \}_{l=1}^{L}$, where $\mathbf{E}_{l,k}^{s}\in \mathbb{R}^{C_{L_{p}}\times h\times w}$. For pyramid feature set of each exemplar, we perform cosine multiplication on each feature pair to generate the similarity matrix:
\begin{equation}\label{eq1}
    \mathbf{S}_{l,k}\left (x^{q},x^{s}\right )= {\rm ReLU}\left (\frac{\mathbf{E}_{l}^{q}\left (x^{q}\right )\cdot \mathbf{E}_{l,k}^{s}\left (x^{s}\right )}{\left \| \mathbf{E}_{l}^{q}\left (x^{q}\right ) \right \|\left \| \mathbf{E}_{l,k}^{s}\left (x^{s}\right ) \right \|}\right ).
\end{equation}
Here, $x^{q}$ and $x^{s}$ denote 2-dimensional spatial positions of query feature map $\mathbf{E}_{l}^{q}$ and exemplar feature map $\mathbf{E}_{l,k}^{s}$, respectively. `$\cdot$' denotes vector dot product. For each similarity tensor, $\mathbf{S}_{l,k}\in \mathbb{R}^{H_{L_{p}}\times W_{L_{p}}\times h\times w}$. 

 As shown in Fig. \ref{fig4}, we concatenate the similarity matrices of the same large layer $L_{p}$ and partition $\left \{\mathbf{S}_{l,k}\right \}_{l=1}^{L}$ accordingly, transforming it into $\left \{\left \{\mathbf{S}_{l,k}\right \}_{l\in L_{p}}\right \}_{p=1}^{P}$ and set it as $\left \{\mathbf{S}_{L_{p},k}\right \}_{p=1}^{P}$. And each tensor in it is $\mathbf{S}_{L_{p},k}\in \mathbb{R}^{\left |L_{p} \right |\times H_{L_{p}}\times W_{L_{p}}\times h\times w}$, where $\left |L_{p} \right |$ is the number of pyramid levels. They are then fed into the Similarity Learning Module (SLM) to produce a learned and fused feature map $\mathbf{M}_k$. Finally, $\mathbf{M}_k$ is input into the regression decoder module to obtain the density map $\mathbf{P}_{k}\in \mathbb{R}^{1\times H\times W}$. The $K$ sets of feature pyramids correspond to the generation of $K$ density maps. The final predicted density map, denoted as $\mathbf{P}$, is obtained by taking the mean of these $K$ maps:
\begin{equation}\label{eq2}
    \mathbf{P}=\frac{\sum_{k=1}^{K}\mathbf{P}_{k}}{K}.
\end{equation}

\subsection{Feature Cross Enhancement}\label{FCE}
The distribution of object features within the query features of the same category is often uneven. Directly matching and counting using the original features can result in varying density values for each object. To address this issue, we propose a Feature Cross Enhancement (FCE) module that aims to bring the object features within the query closer to the exemplar features while also facilitate the exemplar features to be closer to the center position of all object features. By enhancing the proximity of the object features specific to a certain category, the model is able to generate more accurate density values.

In the FCE module, the input features $\mathbf{F}_{l}^{q}$ and $\mathbf{F}_{l}^{s}$ are jointly transformed into $\mathbf{V}^{q}\in \mathbb{R}^{C_{p}^{e}\times H_{L_{p}}\times W_{L_{p}}}$ and $\mathbf{V}^{s}\in \mathbb{R}^{C_{p}^{e}\times K\times h\times w}$ through a convolutional layer. They are then individually passed through other two convolutional layers, with $\mathbf{F}_{l}^{q}$ being transformed into $\mathbf{Q}$ and $\mathbf{F}_{l}^{s}$ into $\mathbf{K}$, which are the same dimensions as $\mathbf{V}^{q}$ and $\mathbf{V}^{s}$. Multiplying the transpose of $\mathbf{Q}$ and $\mathbf{K}$ matrices results in the attention matrix $\mathbf{A}_{l}$:
\begin{equation}\label{eq3}
    \mathbf{A}_{l}= {\rm SoftMax}\left (\mathbf{Q}^{T}\mathbf{K}\right ).
\end{equation}
Then we utilize $\mathbf{A}_{l}$ to separately enhance $\mathbf{F}_{l}^{q}$ and $\mathbf{F}_{l}^{s}$:
\begin{equation}\label{eq4}
\begin{aligned}
    \mathbf{E}_{l}^{q}&= \mathbf{F}_{l}^{q}+{\rm MLP}\left (\mathbf{F}_{l}^{q} \right )\odot {\rm Trans}\left (\mathbf{V}^{s}\mathbf{A}_{l}^{T}\right ) \\
    \mathbf{E}_{l}^{s}&= \mathbf{F}_{l}^{s}+{\rm MLP}\left (\mathbf{F}_{l}^{s} \right )\odot {\rm Trans}\left (\mathbf{V}^{q}\mathbf{A}_{l}\right ).
\end{aligned}
\end{equation}
Here, ${\rm MLP\left (\cdot \right )}$ is a multi-layer perceptron consisting of fully connected layers and activation functions, and used to map the channel vector into a channel-wise feature space of similarity relation. ${\rm Trans\left (\cdot \right )}$ represents the convolutional layer that transforms channel $C_{p}^{e}$ into the original channel $C_{L_{p}}$, and $\odot$ denotes element-wise multiplication.

\subsection{Similarity Learning Module}\label{SLM}
{\bf 4D convolution}. Several existing works \cite{rocco2018neighbourhood,yang2019volumetric,min2021hypercorrelation} have proposed various implementations of 4D convolutions. In our framework, we employ the center-pivot 4D convolution from \cite{min2021hypercorrelation} which sparsifies a significant portion of unimportant weights and computations. This method focuses solely on the information associated with the convolution center, reducing computational overhead while maintaining effectiveness. With 4D convolutions, tensors are fused for each 4D position based on convolution kernel weights, integrating information from the vicinity in 4D space and transforming the vector at that position into the corresponding output dimension.

For the input set of similarity tensors $\left \{\mathbf{S}_{L_{p}}\right \}_{p=1}^{P}$ (here we omit the exemplar subscript $k$), each tensor is fed into its corresponding 4D convolutional module:
\begin{equation}\label{eq5}
    {\mathbf{S}}'_{L_{p}}= f_{L_{p}}^{e}\left ( \mathbf{S}_{L_{p}} \right )\in \mathbb{R}^{C_{out}\times H_{L_{p}}\times W_{L_{p}}\times {h}'\times {w}'},
\end{equation}
where $f_{L_{p}}^{e}\left (\cdot \right )$ is an encoding module composed of multiple 4D convolutional layers, group normalization \cite{wu2018group}, and ReLU activation function. The large strides of the 4D convolution compresses the spatial dimensions $h\times w$ of the exemplar spatial structure to ${h}'\times {w}'$, while embedding the dimensions of all similarity tensors from $\left |L_{p} \right |$ into $C_{out}$.

Next, starting from the apex of pyramid $\left \{{\mathbf{S}}'_{L_{p}}\right \}_{p=1}^{P}$, we proceed to fuse each subsequent layer downwards. For instance, the tensor ${\mathbf{S}}'_{L_{P}}\in \mathbb{R}^{C_{out}\times H_{L_{P}}\times W_{L_{P}}\times {h}'\times {w}'}$ is upsampled on its dimensions $H_{L_{P}}\times W_{L_{P}}$ to match the corresponding dimensions $H_{L_{P-1}}\times W_{L_{P-1}}$ of the layer below. It is then added to the respective tensor ${\mathbf{S}}'_{L_{P-1}}$ of the layer below and passed through a fusion module based on 4D convolution:
\begin{equation}\label{eq6}
    \mathbf{S}_{L_{p-1}}^{Mix}=f_{L_{p-1}}^{Mix}\left ( {\rm upsample}\left ( {\mathbf{S}}'_{L_{p}}\right )+{\mathbf{S}}'_{L_{p-1}} \right ).
\end{equation}
The structure of function $f_{L_{p-1}}^{Mix}$ is identical to that of function $f_{L_{p}}^{e}$, with the difference being that the stride of $f_{L_{p-1}}^{Mix}$ is set to 1, which does not alter the spatial dimensions of the tensor. And the input and output dimensions in $f_{L_{p-1}}^{Mix}$ are all set to $C_{out}$.

$\mathbf{S}_{L_{P-1}}^{Mix}$ is fused with the tensor ${\mathbf{S}}'_{L_{P-2}}$ in a similar manner, iteratively continuing the fusion process with each subsequent layer until reaching the bottom layer of the pyramid ${\mathbf{S}}'_{L_{1}}$. Consequently, we obtain the final fused tensor $\mathbf{S}_{L_{1}}^{Mix}\in \mathbb{R}^{C_{out}\times H_{L_{1}}\times W_{L_{1}}\times {h}'\times {w}'}$. By calculating the mean along the last two dimensions, we derive the fused feature $\mathbf{M}\in \mathbb{R}^{C_{out}\times H_{L_{1}}\times W_{L_{1}}}$.

\subsection{Regression Decoder}\label{RD}
The decoder module used for regression consists of multiple component modules composed of convolutional layers, ReLU activation layers, and upsampling layers. With each component module, the size of feature $\mathbf{M}$ is increased to twice until reaching the size of the input image $H\times W$. Subsequently, it passes through a $1\times 1$ convolutional layer and a ReLU activation layer. The output is the predicted density map.

\subsection{Generalized Loss}\label{GLM}
In previous object counting tasks, ground truth density maps are generated by convolving dot labels with fixed Gaussian kernels. The MSE loss function is then employed for supervised training of the predicted density map. In a recent study \cite{wan2021generalized}, a generalized loss function was proposed that directly measures the distance between the predicted density map and the dot labels. This loss function is based on entropic-regularized unbalanced optimal transport cost.

We represent the predicted results as $\mathbf{A}=\left \{ \left ( a_{i},\mathbf{x}_{i} \right ) \right \}_{i=1}^{n}$, where $a_{i}$ denotes the predicted density value at pixel $\mathbf{x}_{i}\in \mathbb{R}^{2}$. Here, n represents the total number of pixels. Then we denote the predicted density map as $\mathbf{a}=\left [ a_{i} \right ]_{i}$. On the other hand, the ground truth dot label is denoted as $\mathbf{B}=\left \{ \left ( b_{j},\mathbf{y}_{j} \right ) \right \}_{j=1}^{m}$, where $\mathbf{y}_{j}$ indicates the location of the j-th annotation and $b_{j}$ represents the number of objects represented by that annotation. In general, it is assumed that $\mathbf{b}=\left [ b_{j} \right ]_{j}=1_{m}$. The whole loss function can be defined as:
\begin{equation}\label{eq7}
\begin{aligned}
    L_{\mathbf{C}}\left ( \mathbf{A},\mathbf{B} \right )=&\min_{\mathbf{D}}\left \langle \mathbf{C},\mathbf{D} \right \rangle-\varepsilon H\left ( \mathbf{D} \right )+\tau \left \| \mathbf{D}1_{m}-\mathbf{a} \right \|_{2}^{2}\\
    &+\tau \left \| \mathbf{D}^{T}1_{n}-\mathbf{b} \right \|_{1},
\end{aligned}
\end{equation}
where $\mathbf{C}$ is the transport cost matrix, with $C_{ij}$ representing the cost of moving the predicted density at $\mathbf{x}_{i}$ to the ground truth dot annotation $\mathbf{y}_{j}$. $\mathbf{D}$ is the transport matrix that assigns fractional weights to associate each location $\mathbf{x}_{i}$ in $\mathbf{A}$ with its corresponding $\mathbf{y}_{j}$ in $\mathbf{B}$ for cost calculation. The optimal transport cost is obtained by minimizing the loss over $\mathbf{D}$. $H\left ( \mathbf{D} \right )=-\sum _{ij}D_{ij}{\rm log}D_{ij}$ is the entropic regularization. The intermediate density map representation $\hat{\mathbf{a}}=\mathbf{D}1_{m}$ is constructed from the ground truth annotations, while $\hat{\mathbf{b}}=\mathbf{D}^{T}1_{n}$ is the reconstruction of the ground truth dot annotations.

\subsection{Dynamic Image Scale}\label{DIS}
Certain sample images may contain objects that are small sizes or densely distributed, leading to challenges in effectively distinguishing between individual objects. This results in overlapping density within the predicted density map, thereby impacting model performance. To address this issue, we dynamically resize the input images based on the size of exemplar boxes before entering the backbone. This resizing is performed proportionally to the dimensions of the exemplar boxes, allowing the model to better recognize samples containing smaller objects. For an input image $\mathbf{X}$, we compute the average length and width of K exemplar boxes $\mathbf{B}\in \mathbb{R}^{K\times 2}$:
\begin{equation}\label{eq8}
    \bar{\mathbf{B}}=\frac{\sum_{k=1}^{K}\mathbf{B}_{k}}{K}.
\end{equation}
If ${\rm min}\left (\bar{\mathbf{B}}\right )$ is below a threshold $\gamma$ , we calculate the scale of image expansion:
\begin{equation}\label{eq9}
    scale=\frac{\gamma - {\rm min}\left (\bar{\mathbf{B}}\right )}{\eta }+1,
\end{equation}
where both $\gamma$ and $\eta$ are hyperparameters to be tuned. Finally, the image size and exemplar boxes $\mathbf{B} $ are simultaneously expanded by the determined scale value before being input into the model.

\begin{table*}[t]
\centering
\setlength{\tabcolsep}{2.5mm}{
    \begin{tabular}{@{}lc|cccc|cccc}
    \toprule
		\multirow{3}{*}{Methods} &\multirow{3}{*}{Backbone} & \multicolumn{4}{c}{3shot} &\multicolumn{4}{c}{1shot} \\ \cmidrule(lr){3-6}\cmidrule(lr){7-10} && \multicolumn{2}{c}{Val}  & \multicolumn{2}{c}{Test} & \multicolumn{2}{c}{Val}  & \multicolumn{2}{c}{Test}\\
      && MAE & RMSE   &  MAE  & RMSE &MAE & RMSE   &  MAE  & RMSE\\
				\midrule
		GMN \cite{lu2019class} &ResNet-50 &29.66&89.81&26.52&124.57 &--- &--- &--- &--- \\
	    MAML \cite{finn2017model} &ConvNet &25.54&79.44&24.90&112.68 &--- &--- &--- &--- \\
	    FamNet \cite{ranjan2021learning} &ResNet-50 &23.75&69.07&22.08&99.54 &26.55&77.01&26.76&110.95 \\
        CFOCNet \cite{yang2021class} &ResNet-50 &21.19&61.41&22.10&112.71 &27.82&71.99&28.60&123.96  \\
        LaoNet \cite{lin2021object} &VGG-19 &--- &--- &--- &--- &17.11&56.81&15.78&97.15 \\
        BMNet+ \cite{shi2022represent} &ResNet-50 &15.74&58.53&14.62&91.83 &17.89&61.12&16.89&96.65 \\
        SAFECount \cite{you2023few} &ResNet-18 &15.28&47.20&14.32&85.54 &--- &--- &--- &--- \\
        SPDCN \cite{Lin_2022_BMVC} &VGG-19 &14.59&49.97&13.51&96.80 &--- &--- &--- &--- \\
        CounTR \cite{liu2022countr} &ViT/ConvNet &13.13&49.83&11.95&91.23 &13.15&49.72&12.06&90.01 \\
        LOCA \cite{djukic2023low} &ResNet-50 &10.24&32.56&10.79&\textbf{56.97} &11.36&38.04&12.53&75.32 \\
        \textbf{SSD(ours)} &ResNet-50 &\textbf{9.73}&\textbf{29.72}&\textbf{9.58}&64.13
        &\textbf{11.03}&\textbf{34.83}&\textbf{11.61}&\textbf{71.55}\\
	    	\bottomrule
    \end{tabular}}
    \caption{Comparison with state-of-the-art approaches on the FSC-147 dataset. `---' means the result is not reported.}
	\label{tab1}
\end{table*}

\section{Experiments}\label{sec:exp}
\subsection{Datasets and Metrics}\label{MAD}

{\bf Datasets}. {\bf FSC-147} is a comprehensive multi-class few-shot object counting dataset. It comprises a total of 6,135 images covering 89 distinct object categories. The images in the dataset exhibit significant variations in terms of object counts, ranging from as low as 7 to as high as 3,731 objects, with an average count of 56 per image. Notably, each image in the dataset is accompanied by three or four exemplar images that are annotated with bounding boxes. To facilitate experimentation, the dataset is further divided into training, validation, and testing subsets, with each subset containing 29 non-overlapping object categories.

{\bf CARPK} is a class-specific car counting dataset, which consists of 1448 images of parking lots from a bird’s view. These images are captured from four different parking lots, encompassing various scenes. The training set comprises three scenes, while a separate scene is designated for test.

{\bf Metrics}. We employ Mean Average Error (MAE) and Root Mean Squared Error (RMSE) as performance metrics for evaluating the SSD method, as these metrics are widely utilized in counting tasks.
\begin{equation}\label{eq10}
\begin{aligned}
MAE&=\frac{1}{N}\sum_{i=1}^{N}\left | C_{pred}^{i}-C^{i} \right |,\\
RMSE&=\sqrt{\frac{1}{N}\sum_{i=1}^{N}(C_{pred}^{i}-C^{i})^{2}},
\end{aligned}
\end{equation}
where $N$ is the number of all the query images, $C^{i}$ and $C_{pred}^{i}$ are the ground truth and the predicted number of objects for $i$-th image respectively.

\begin{figure}[t]
\centering
{\includegraphics[width=0.48\textwidth]{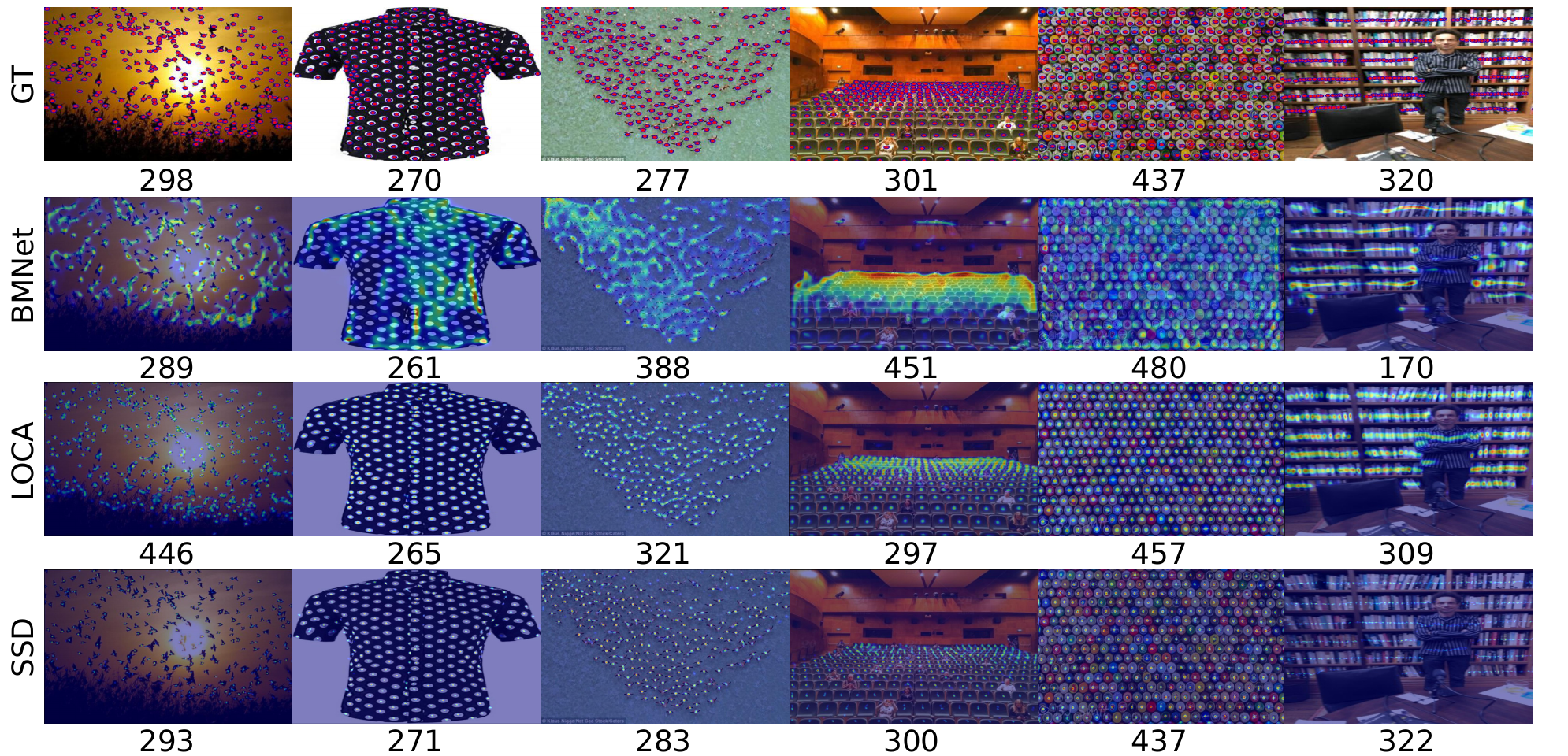}}
\caption{Qualitative results on the FSC-147 dataset.}
   \label{fig5}
\end{figure}

\subsection{Implementation Details}\label{Details}

{\bf Architecture Details}. Our approach involves resizing the input image initially to $384\times 576$. Then the image is dynamically resized to the suitable scale, followed by the application of the pre-trained ResNet-50 backbone, utilizing features extracted from the final three layers. The number of features at each layer, denoted as $\left | L_{p} \right |$, is 4, 6, and 3, with corresponding feature channel dimensions of 512, 1024, and 2048, respectively. The size of exemplar features extracted from each layer is uniformly resized to $16\times 16$. In the FCE module, the embedded channel dimension is set to half of dimension of the input feature. In the Similarity Learning module, the 4D convolution module consists of three component modules with output dimensions of 32, 128, and 256, respectively. To fuse the three layers of similarity tensors, two fusion modules are required, each containing three component modules, and all the output dimensions are 256. $\gamma$ and $\eta$ in DIS method are set to 32 and 12.

{\bf Training Details}. We apply AdamW \cite{loshchilov2017decoupled} as the optimizer with a learning rate of $1\times10^{-4}$ and the learning rate decays with a rate of 0.95 after each epoch. The batch size is 4 and the model is trained for 100 epochs.

\subsection{Comparison with State of the Art}\label{compare with sota}
We evaluate the proposed SSD on the FSC-147 dataset with several state-of-the-art methods. the results are summarized in Tab.\ref{tab1}. We conduct both 3-shot and 1-shot experiments on the dataset. SSD consistently outperforms existing methods in terms of performance.

In the 3-shot scenario, even compared to the recent state-of-the-art method LOCA \cite{djukic2023low}, SSD demonstrates superior performance on the val set with a 5.0\% improvement in MAE and an 8.7\% improvement in RMSE. Notably, SSD also exhibits better performance on the test set with a 11.2\% improvement in MAE. 

In the 1-shot scenario, SSD surpasses all previous state-of-the-art methods. This underscores the minimal dependence of SSD on reference samples, showcasing its robust adaptability to scenarios with limited available data.

{\bf Qualitative Results}. In Fig.\ref{fig5}, we visualize and compare the predicted density maps of BMNet, LOCA, and SSD. The results demonstrate that SSD has higher accuracy compared to the other two methods.

\begin{table}[t]
\centering
\setlength{\tabcolsep}{1mm}{
    \begin{tabular}{@{}lccc}
    \toprule
		\multirow{2}{*}{Methods} & \multirow{2}{*}{BackBone}  & \multicolumn{2}{c}{Test} \\
         &&  MAE  & RMSE\\
				\midrule
	    FamNet \cite{ranjan2021learning} &ResNet-50 &28.84&44.47 \\
        BMNet \cite{shi2022represent} &ResNet-50 &10.44&13.77 \\
        LOCA \cite{djukic2023low} &ResNet-50 &9.97&12.51 \\
        \textbf{SSD(ours)} &ResNet-50 &\textbf{9.58}&\textbf{12.15} \\
	    	\bottomrule
    \end{tabular}}
    \caption{Comparison with the state-of-the-art approaches on the CARPK dataset. }
	\label{tab2}
\end{table}

\subsection{Cross-dataset Generalization}\label{CdG}
Following \cite{ranjan2021learning}, we validate the generalizability of SSD on the CARPK dataset. The model is trained on the FSC-147 dataset and then tested on the CARPK dataset, with the car category samples excluded during training. During testing, we randomly select twelve annotations from the CARPK dataset as exemplars to count cars in images. The experimental results is presented in Tab.\ref{tab2}. SSD outperforms three other methods, achieving an improvement of 3.9\% in MAE and 2.9\% in RMSE compared to the most recent state-of-the-art method LOCA.

\begin{table}[t]
\centering
\resizebox{\linewidth}{!}{
\begin{tabular}{@{}ccccccc}
\toprule
\multirow{2}{*}{FCE} &\multirow{2}{*}{G-Loss} &\multirow{2}{*}{DIS} & \multicolumn{2}{c}{Val}  & \multicolumn{2}{c}{Test} \\
&&& MAE & RMSE   &  MAE  & RMSE\\
\midrule
\ding{55} &\ding{55} &\ding{55} &18.96&64.44&16.75&108.12 \\
\checkmark &\ding{55} &\ding{55} &18.56&61.69&16.39&108.64 \\
\ding{55} &\checkmark &\ding{55} &15.14&54.75&14.99&107.40 \\
\ding{55} &\ding{55} &\checkmark &13.68&45.53&14.25&91.17 \\
\checkmark &\checkmark &\ding{55} &13.92&51.08&14.43&106.84 \\ 
\ding{55} &\checkmark &\checkmark &10.50&31.86&11.38&74.45 \\
\checkmark &\ding{55} &\checkmark &13.37&39.26&12.90&82.16 \\
\checkmark &\checkmark &\checkmark &9.73&29.72&9.58&64.13 \\
\bottomrule
\end{tabular}}
\caption{Ablation studies on the FSC-147 dataset. `G-Loss' means Generalized Loss. `DIS' denotes Dynamic Image Scale.}
\label{tab3}
\end{table}

\subsection{Ablation Study}\label{ablation}
We design a series of experiments to validate the individual contributions of the FCE module, generalized loss, and dynamic image scale on the performance improvement. In the absence of the FCE module, the model directly computes the similatity between $\mathbf{F}_{l}^{q}$ and $\mathbf{F}_{l}^{s}$. When excluding the generalized loss, we replace it with the more commonly used MSE loss. Each component undergoes four sets of comparative experiments with and without that component. 

{\bf FCE module}. Analysis of the four sets of experiments involving the FCE module reveals a consistent improvement in model performance. The addition of the FCE module results in a performance boost ranging from 2\% to 19\% in MAE and up to 14\% in RMSE. This indicates that FCE module significantly enhances the ability of model to recognize objects within a given category by minimizing the distance between individual object features and exemplar features, leading to improved accuracy and uniformity in similarity and final density predictions across object positions.

{\bf Generalization loss}. The contribution of generalization loss is pronounced to performance improvement. The four sets of comparative experiments show performance gains ranging from 10\% to 27\% in MAE and 1\% to 30\% in RMSE. The substantial improvement attributed to the generalization loss is due to the more precise recognition capabilities compared to MSELoss. By measuring point-to-point distance loss between predicted values and ground true labels, the generalized loss effectively guides the model to accurately locate object center positions.

{\bf Dynamic image scale}. The utilization of dynamic image scale also significantly improves model performance, particularly for dense samples. Expanding image scales proves effective in distinguishing between individual objects and counting them separately. Application of this method results in performance improvements ranging from 15\% to 35\% in MAE and 14\% to 43\% in RMSE.

\begin{figure}[t]
\centering
\subfigure[Channel ratio in FCE module]
{\includegraphics[width=0.227\textwidth]{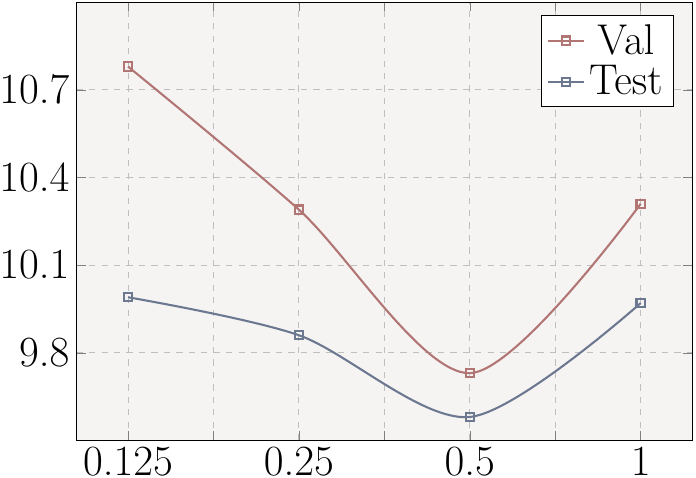}} \quad
\subfigure[Number of component modules in fusion module]
{\includegraphics[width=0.23\textwidth]{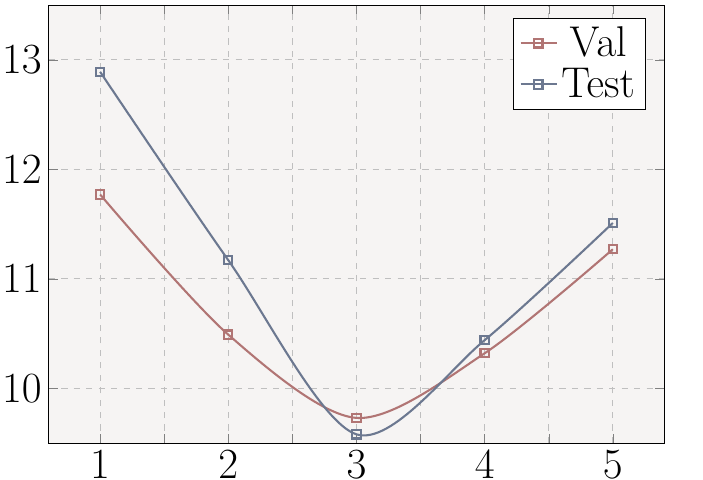}}
\caption{Ablation study of channel ratio in FCE module and number of component modules in fusion module. The vertical coordinates are the values of MAE on the val set.}
   \label{fig6}
\end{figure}

{\bf Channel ratio in the FCE module}. The query features and example features are embedded into another channel before they enhance each other. We conduct a series of experiments to determine the optimal ratio of the embedded channel length to the original channel length, setting various ratios at $\frac{1}{8}$, $\frac{1}{4}$, $\frac{1}{2}$, and 1. We then train the model to achieve the best performance and present the experimental results in Fig.\ref{fig6} (a). It is observed that as the ratio increases, the model performance peaks at a ratio of $\frac{1}{2}$ and then begins to deteriorate.

{\bf Number of component modules in fusion module}. The fusion module used to integrate tensors from different levels of the similarity pyramid is composed of several component modules. We set the number of component modules to range from 1 to 5 and conduct experiments on the FSC-147 dataset, with the results displayed in Fig.\ref{fig6} (b). The line graph in the figure indicates that as the number of component modules increases from 1 to 3, the performance of the model gradually improves, peaking at 3, and then begins to decline. This decline could be attributed to an increase in the number of model parameters due to more component modules, leading to overfitting and negatively affecting model performance.

\section{Conclusion}\label{conclusion}
We propose a novel few-shot object counting method, SSD, which leverages a point-to-point 4D space to learn the spatial similarity distribution between query and exemplar features. In contrast to existing methods, we exploit the distribution information of similarity, enabling accurate identification of the position and precise prediction of the count for objects of arbitrary classes. Additionally, we introduce a Feature Cross Enhancement (FCE) module that enhances the interaction between query and exemplar features, reducing the feature distance within the same class for improved matching. Experimental results on datasets such as FSC-147 and CARPK demonstrate that SSD outperforms state-of-the-art methods.

\section*{Acknowledgments}

This work was supported in part by National Natural Science Foundation of China under the Grants 62371235 and 62072246, and in part by Key Research and Development Plan of Jiangsu Province (Industry Foresight and Key Core Technology Project) under the Grant BE2023008-2.

\bibliographystyle{named}
\bibliography{ijcai24}

\end{document}